\def\year{2020}\relax
\documentclass[letterpaper]{article} 
\usepackage{aaai20}  
\usepackage{times}  
\usepackage{helvet} 
\usepackage{courier}  
\usepackage[hyphens]{url}  
\usepackage{graphicx} 
\usepackage{graphicx}  
\nocopyright
\usepackage{latexsym}
\usepackage{xspace}
\usepackage{xcolor}
\usepackage{amsmath}
\usepackage{amsfonts}       
\usepackage{nicefrac}       
\usepackage{microtype}  
\usepackage{amssymb}
\usepackage{stmaryrd}
\usepackage{array}
\usepackage{graphicx}
\usepackage{cases}
\usepackage{watermark}
\usepackage{pbox}
\usepackage{soul}
\usepackage{enumitem}
\usepackage{makecell}
\usepackage[draft]{todonotes}
\usepackage[ruled,noline,linesnumbered]{algorithm2e}
\usepackage{enumitem}
\usepackage{boxedminipage}
\usepackage{lipsum}
\usepackage{arydshln}
\usepackage{figsize}
\usepackage{multirow}
\usepackage{booktabs}
\usepackage{pifont}

\newcommand{\newcite}[1]{\citeauthor{#1} \shortcite{#1}}
\newcommand{\dataset}{\textsc{WinoGrande}\xspace}
\newcommand{\datasetdeb}{\textsc{WinoGrande}$_\textit{debiased}$\xspace}
\newcommand{\datasetall}{\textsc{WinoGrande}$_\textit{all}$\xspace}
\newcommand{\algoname}{\textsc{AfLite}\xspace}

\newcommand{\robertaembed}{RoBERTa$_\text{embed}$}
\newcommand{\blank}{\underline{\ \ \ \ \ \ }\xspace}

\newcommand{\cmark}{{\color{green}\ding{51}}\xspace}%
\newcommand{\xmark}{{\color{red}\ding{55}}\xspace}%

\newcommand{\para}[1]{\vspace{2mm}\noindent\textbf{#1} \xspace \xspace}%

\definecolor{emerald}{rgb}{0.31, 0.78, 0.47}
\definecolor{piggypink}{rgb}{0.99, 0.87, 0.9}
\definecolor{teagreen}{rgb}{0.82, 0.94, 0.75}

\newcommand{\allenai}{$^{*}$}
\newcommand{\uw}{$^{\dagger}$}

\title{\dataset: An Adversarial Winograd Schema Challenge at Scale}

\author{
Keisuke Sakaguchi\allenai, Ronan Le Bras\allenai, Chandra Bhagavatula\allenai, Yejin Choi\allenai\uw\\
\allenai Allen Institute for Artificial Intelligence \uw University of Washington \\
  {\tt \{keisukes, ronanlb, chandrab, yejinc\}@allenai.org}
}

\begin{document}

\maketitle

\begin{abstract}
The Winograd Schema Challenge (WSC) \cite{winograd},
a benchmark for commonsense reasoning,  
is a set of $273$ expert-crafted pronoun resolution problems originally designed to be unsolvable for statistical models that rely on selectional preferences or word associations.
However, recent advances in neural language models have already reached around 90\% accuracy on variants of WSC.   
This raises an important question whether these models have truly acquired robust commonsense capabilities or whether they rely on spurious biases in the datasets that lead to an overestimation of the true capabilities of machine commonsense.

To investigate this question, we introduce \textbf{\dataset}, a large-scale dataset of 44k problems, inspired by the original WSC design, but adjusted to improve both the scale and the hardness of the dataset. The key steps of the dataset construction consist of (1) a carefully designed crowdsourcing procedure, followed by (2) systematic bias reduction using a novel \textsc{AfLite} algorithm that generalizes human-detectable \emph{word associations} to machine-detectable \emph{embedding associations}.
The best state-of-the-art methods on \dataset achieve 59.4 -- 79.1\%, which are $\sim$15-35\% (absolute) below human performance of 94.0\%, depending on the amount of the training data allowed (2\% -- 100\% respectively).

Furthermore, we establish new state-of-the-art results on \emph{five} related benchmarks --- 
WSC ($\rightarrow$ \textbf{90.1\%}),
DPR ($\rightarrow$ {\textbf{93.1\%}}),
COPA($\rightarrow$ \textbf{90.6\%}),
KnowRef ($\rightarrow$ \textbf{85.6\%}),
and Winogender ($\rightarrow$ \textbf{97.1\%}).
These results have dual implications: on one hand, they  demonstrate the effectiveness of \dataset when used as a resource for transfer learning. On the other hand, they raise a concern that we are likely to be overestimating the true capabilities of machine commonsense across all these benchmarks.
We emphasize the importance of algorithmic bias reduction in existing and future benchmarks to mitigate such overestimation.
\end{abstract}

\label{sec:introduction}
\section{Introduction}


The Winograd Schema Challenge (WSC) \cite{winograd}, proposed as an alternative to the Turing Test~\cite{turing1950computing}, has been used as a benchmark for evaluating commonsense reasoning. 
WSC are designed to be pronoun resolution problems (see examples in Table \ref{tab:wsc-examples}) that are trivial for humans but hard for machines that merely rely on statistical patterns without true capabilities of commonsense reasoning.  
However, recent advances in neural language models have already reported around 90\% accuracy on a variant of WSC dataset.\footnote{\url{https://github.com/pytorch/fairseq/tree/master/examples/roberta}. We note that this variant aggregates the original WSC, PDP~\cite{morgenstern2016planning} and additional PDP-style examples, and recasts them into True/False binary problems.}
This raises an important question:
\begin{itemize}
\item[] \textit{Have neural language models successfully acquired commonsense or are we overestimating the true capabilities of machine commonsense}?
\end{itemize}
This question about the potential overestimation leads to another crucial question regarding potential unwanted biases that the large-scale neural language models might be exploiting, essentially solving the problems \emph{right}, but for \emph{wrong} reasons.
While WSC questions are expert-crafted,  recent  studies  have  shown that they are nevertheless prone to incidental biases.  \newcite{trichelair2018evaluation} have reported \textit{word-association} (13.5\% of the cases, see Table~\ref{tab:wsc-examples} for examples) as well as other types of \textit{dataset-specific} biases.
While such biases and annotation artifacts are not apparent for individual instances, they get introduced in the dataset as problem authors subconsciously repeat similar problem-crafting strategies.

\begin{table*}[t]
\centering
\begin{tabular}{clll}
\hline
\multicolumn{3}{c}{\small{Twin sentences }}      & \small{Options (\textbf{answer})} \\ \hline
\multirow{2}{*}{\small{\cmark(1)}}   & \small{a} & \small{The trophy doesn't fit into the brown suitcase because \textbf{it}'s too \underline{\textit{large}}.} & \small{\textbf{trophy} / suitcase} \\
                             & \small{b} & \small{The trophy doesn't fit into the brown suitcase because \textbf{it}'s too \underline{\textit{small}}.} & \small{trophy / \textbf{suitcase}} \\ \hline
\multirow{2}{*}{\small{\cmark(2)}}  & \small{a} & \small{Ann asked Mary what time the library closes, \underline{\textit{because}} \textbf{she} had forgotten.} & \small{\textbf{Ann} / Mary} \\
                             & \small{b} & \small{Ann asked Mary what time the library closes, \underline{\textit{but}} \textbf{she} had forgotten.} & \small{Ann / \textbf{Mary}}          \\ \hline
\multirow{2}{*}{\small{\xmark(3)}} & \small{a} & \small{The tree fell down and crashed through the roof of my house. Now, I have to get \textbf{it} \underline{\textit{removed}}.} &  \small{\textbf{tree} / roof} \\
                             & \small{b} & \small{The tree fell down and crashed through the roof of my house. Now, I have to get \textbf{it} \underline{\textit{repaired}}.} &  \small{tree / \textbf{roof}} \\ \hline                     
\multirow{2}{*}{\small{\xmark(4)}} & \small{a} & \small{The lions ate the zebras because \textbf{they} are \underline{\textit{predators}}.} &  \small{\textbf{lions} / zebras} \\
                             & \small{b} &  \small{The lions ate the zebras because \textbf{they} are \underline{\textit{meaty}}.}& \small{lions / \textbf{zebras}} \\ \hline                     
\end{tabular}
\caption{WSC problems are constructed as pairs (called \textit{twin}) of nearly identical questions with two answer choices. The questions include a \underline{\textit{trigger word}} that flips the correct answer choice between the questions. Examples (1)-(3) are drawn from WSC~\cite{winograd} and (4) from DPR~\cite{rahman-ng-2012-resolving}). Examples marked with \xmark have language-based bias that current language models can easily detect. Example (4) is undesirable since the word ``predators" is more often associated with the word ``lions", compared to ``zebras"}
\label{tab:wsc-examples}
\end{table*}

To investigate this question about the true estimation of the machine commonsense capabilities, we introduce \textbf{\dataset}, a new dataset with 44k problems that are inspired by the original design of WSC, but modified to improve both the scale and hardness of the problems. The key steps in \dataset construction consist of (1) a carefully designed crowdsourcing procedure, followed by (2) a novel algorithm \algoname that
generalizes human-detectable biases based on \emph{word} occurrences to machine-detectable biases based on \emph{embedding} occurrences.
The key motivation of our approach is that it is difficult for humans to write problems without accidentally inserting unwanted biases. 

While humans find \dataset problems trivial with 94\% accuracy, best state-of-the-art results, including those from RoBERTa \cite{Liu2019RoBERTaAR} are considerably lower, ranging between 59.4\% - 79.1\% depending on the amount of training data provided (from 800 to 41k instances), which falls 15 - 35\% (absolute) below the human-level performance.

Furthermore, we also demonstrate that \dataset provides transfer learning to other existing WSC and related benchmarks, achieving new SOTA performances on \emph{five} of them, including 
the original WSC~\cite{winograd} ($\rightarrow$ \textbf{90.1\%}),
DPR~\cite{rahman-ng-2012-resolving} ($\rightarrow$ {\textbf{93.1\%}}),
COPA~\cite{copa} ($\rightarrow$ \textbf{90.6\%}),
KnowRef~\cite{emami-etal-2019-knowref} ($\rightarrow$ \textbf{85.6\%}),
and Winogender~\cite{rudinger-etal-2018-gender} ($\rightarrow$ \textbf{97.1\%}).

Although the improvements of SOTA over multiple challenging benchmarks are exciting, we cautiously note that these positive results must be taken with a grain of salt.
The result might also indicate the extent to which spurious effects are prevalent in existing datasets, which runs the risk of overestimating the true capabilities of machine intelligence on commonsense reasoning.
More generally, human-crafted problems and tasks (regardless of whether they are crowdsourced or by experts) contains annotation artifacts in many cases, and algorithmic bias reduction such as \algoname is essential to mitigate such dataset-specific bias.

\section{Crowdsourcing \dataset at Scale}
\label{sec:collection}
WSC problems have been considered challenging to craft by crowdsourcing due to the structural constraints of twins and the requirement of linguistic knowledge (Table~\ref{tab:wsc-examples}). Nevertheless, we present an effective approach to creating a large-scale dataset (\dataset) of WSC problems while maintaining its original properties -- i.e. trivial for humans but hard for AI systems. Our approach consists of a carefully designed crowdsourcing task followed by a novel adversarial filtering algorithm (\S\ref{sec:filtering}) that systematically removes biases in the data.

\para{Enhancing Crowd Creativity}
Creating twin sentences from scratch puts a high cognitive load on crowd workers who thereby subconsciously resort to writing pairs that are lexically and stylistically repetitive.
To encourage creativity and reduce their cognitive load, we employed \textit{creativity from constraints}~\cite{stokes2005creativity} -- a psychological notion which suggests that appropriate constraints can help structure and drive creativity. In practice, crowd workers are primed by a randomly chosen topic as a suggestive context (details below), while they are asked to follow precise guidelines on the structure of the curated data.

\para{Crowdsourcing Task}
We collect \dataset problems via crowdsourcing on Amazon Mechanical Turk (AMT).\footnote{Our datasets, crowdsourcing interface, and models are available at \url{http://winogrande.allenai.org}.}
Workers are asked to write twins sentences (as shown in Table~\ref{tab:wsc-examples}) that meet the requirements for WSC problems (e.g., avoiding word association, non-zero but small edit distance).
To avoid repeating the same topics, workers were instructed to randomly pick an \textit{anchor} word(s) from a randomly assigned WikiHow article\footnote{\url{https://www.wikihow.com/Special:Randomizer}} and to ensure that the twin sentences contain the \textit{anchor} word.
The \textit{anchor} word does not have to be a \textit{trigger} word, but we ensured that it is not a function word such as \textit{the, it, he, of}.
In our pilot experiments, we found that this constraint drastically improves worker's creativity and diversity of topics.
Additionally, workers were instructed to keep twin sentence length in between $15$ and $30$ words while maintaining at least $70\%$ word overlap between a pair of twins.\footnote{The workers met minimum qualification in AMT: $99\%$ approval rate, $5k$ approvals.
The reward was $\$0.4$ per twin sentences.}
Following the original WSC problems, we aimed to collect twins in two different domains -- (i) social commonsense: a situation involving two same gender people with contrasting attributes, emotions, social roles, etc., and (ii) physical commonsense: a context involving two physical objects with contrasting properties, usage, locations, etc.
In total, we collected 77k questions (i.e., 38k twins).

\para{Data Validation}
We validate each collected question through a distinct set of three crowd workers. Following a rigorous process, a question is deemed valid if (1) the majority of the three workers chooses the correct answer option, (2) they agree that the two answer options are unambiguous (one option is clearly more plausible than the other) and (3) the question cannot be answered simply by word association in which local context around the target pronoun is given (e.g., ``because \textbf{it} was going so fast.'' (\textbf{race car} / school bus)).\footnote{For each sentence validation, workers were paid \$0.03.}
As a result, 68\% of the questions (53k) were deemed valid and we discarded the invalid questions.

While our crowdsourcing procedure addresses some amount of instance-level biases like word association, it is still possible that the constructed dataset has \textit{dataset-specific} biases -- especially after it has been scaled up. To address this challenge, we propose a method for systematic bias reduction.

\section{Algorithmic Data Bias Reduction}
\label{sec:filtering}
Several recent studies \cite{gururangan-etal-2018-annotation,poliak-etal-2018-hypothesis,tsuchiya-2018-performance,niven-kao-2019-probing,geva2019we}
have reported the presence of \textit{annotation artifacts} in large-scale datasets.
Annotation artifacts are unintentional patterns in the data that leak information about the target label in an undesired way.
State-of-the-art neural models are highly effective at exploiting such artifacts to solve problems \emph{correctly}, but for \emph{incorrect} reasons. 
To tackle this persistent challenge with dataset biases, we propose \algoname{} -- a novel algorithm that can systematically reduce biases using state-of-the-art contextual representation of words.

\para{Light-weight adversarial filtering}
Our approach builds upon the adversarial filtering (AF) algorithm proposed by \newcite{Zellers2018SWAGAL}, but makes two key improvements: (1) \algoname{} is much more broadly applicable (by not requiring over generation of data instances) and (2) it is considerably more lightweight (not requiring re-training a model at each iteration of AF).
Overgenerating machine text from a language model to use in test instances runs the risk of distributional bias where a discriminator can learn to distinguish between machine generated instances and human-generated ones. In addition, AF depends on training a model at each iteration, which comes at extremely high computation cost when being adversarial to a model like BERT~\cite{bert}.\footnote{\algoname{} is designed for filtering instances so that the resulting dataset is less biased, whereas the original AF algorithm \cite{Zellers2018SWAGAL} is designed for ``generating and modifying'' individual instances, such as by creating better distractors. \algoname{} and AF are therefore different in their goals and hence difficult to compare directly.}

Instead of manually identified lexical features, we adopt a dense representation of instances using their \emph{pre-computed} neural network embeddings. In this work, we use RoBERTa~\cite{Liu2019RoBERTaAR} fine-tuned on a small subset of the dataset.
Concretely, we use 6k instances (5k for training and 1k for validation) from the dataset (containing 53k instances in total) to fine-tune RoBERTa (referred to as \robertaembed{}).
We use \robertaembed{} to pre-compute the embeddings for the rest of the instances (47k) as the input for \algoname{}.
We discard the 6k instances from the final dataset.

Next, we use an ensemble of linear classifiers (logistic regressions) trained on random subsets of the data to determine whether the representation used in \robertaembed{} is strongly indicative of the correct answer option. If so, we discard the corresponding instances and proceed iteratively.

\begin{algorithm}[t]
\small
\DontPrintSemicolon
\SetAlgoNoEnd
\KwIn{dataset $\mathcal{D}=({\bf X}, {\bf y})$, ensemble size $n$, training set size $m$, cutoff size $k$, filtering threshold $\tau$}
\KwOut{dataset $\mathcal{D}'$}
$\mathcal{D}'=\mathcal{D}$\\
\While{$|\mathcal{D}'| > m$}{
    \tcp{Filtering phase}
    \ForAll{$e \in \mathcal{D}'$}{
        Initialize the ensemble predictions $E(e) = \emptyset$\\
    }
    \For{iteration $i: 1..{n}$}{
        Random partition $(\mathcal{T}_i, \mathcal{V}_i$) of $\mathcal{D}'$  s.t. $|\mathcal{T}_i|=m$\\
        Train a linear classifier $\mathcal{L}$ on $\mathcal{T}_i$\\
        \ForAll{$e=({\bf x}, y) \in \mathcal{V}_i$}{
            Add $\mathcal{L}({\bf x})$ to $E(e)$\\
        }
    }
    \ForAll{$e=({\bf x}, y) \in \mathcal{D}'$}{
        $score(e) = \frac{|\{p \in E(e)~\textit{s.t.}~p = y\}|}{|E(e)|}$\\
    }
    Select the top-$k$ elements $\mathcal{S}$ in $\mathcal{D}'$ s.t. $score(e) \geq \tau$\\
    $\mathcal{D}' = \mathcal{D}' \setminus \mathcal{S}$\\
    \If{$|\mathcal{S}| < k$}{
        \textbf{break}
    }
}
\Return $\mathcal{D}'$
\caption{\algoname}
\label{alg:filter}
\end{algorithm}

Algorithm~\ref{alg:filter} provides the implementation of \algoname{}.
The algorithm takes as input the \emph{pre-computed} embeddings ${\bf X}$ and labels ${\bf y}$, along with the size $n$ of the ensemble, the training size $m$ for the classifiers in the ensemble, the size $k$ of the filtering cutoff, and the filtering threshold $\tau$.
At each filtering phase, we train $n$ linear classifiers on different random partitions of the data and we collect their predictions on their corresponding validation set. For each instance, we compute its \emph{score} as the ratio of correct predictions over the total number of predictions. We rank the instances according to their score and remove the top-$k$ instances whose score is above threshold $\tau$. We repeat this process until we remove fewer than $k$ instances in a filtering phase or there are fewer than $m$ remaining instances. When applying \algoname{} to \dataset{}, we set $m=10,000$, $n=64$, $k=500$, and $\tau=0.75$.

This approach is also reminiscent of recent work in NLP on adversarial learning \cite{Chen2018MultinomialAN,belinkov-etal-2019-adv-removal,elazar-goldberg-2018-adversarial}. \newcite{belinkov-etal-2019-adv-removal} propose an adversarial removal technique for NLI which encourages models to learn representations that are free of hypothesis-only biases.
When proposing a new benchmark, however, we cannot enforce that any future model will purposefully avoid learning spurious correlations in the data. In addition, while the hypothesis-only bias is an insightful bias in NLI, we make no assumption about the possible sources of bias in \dataset. Instead, we adopt a more proactive form of bias reduction by relying on state-of-the-art (statistical) methods to uncover undesirable dataset shortcuts.

\begin{figure*}[t]
  \centering
  \SetFigLayout{2}{4}
    \subfigure{\includegraphics{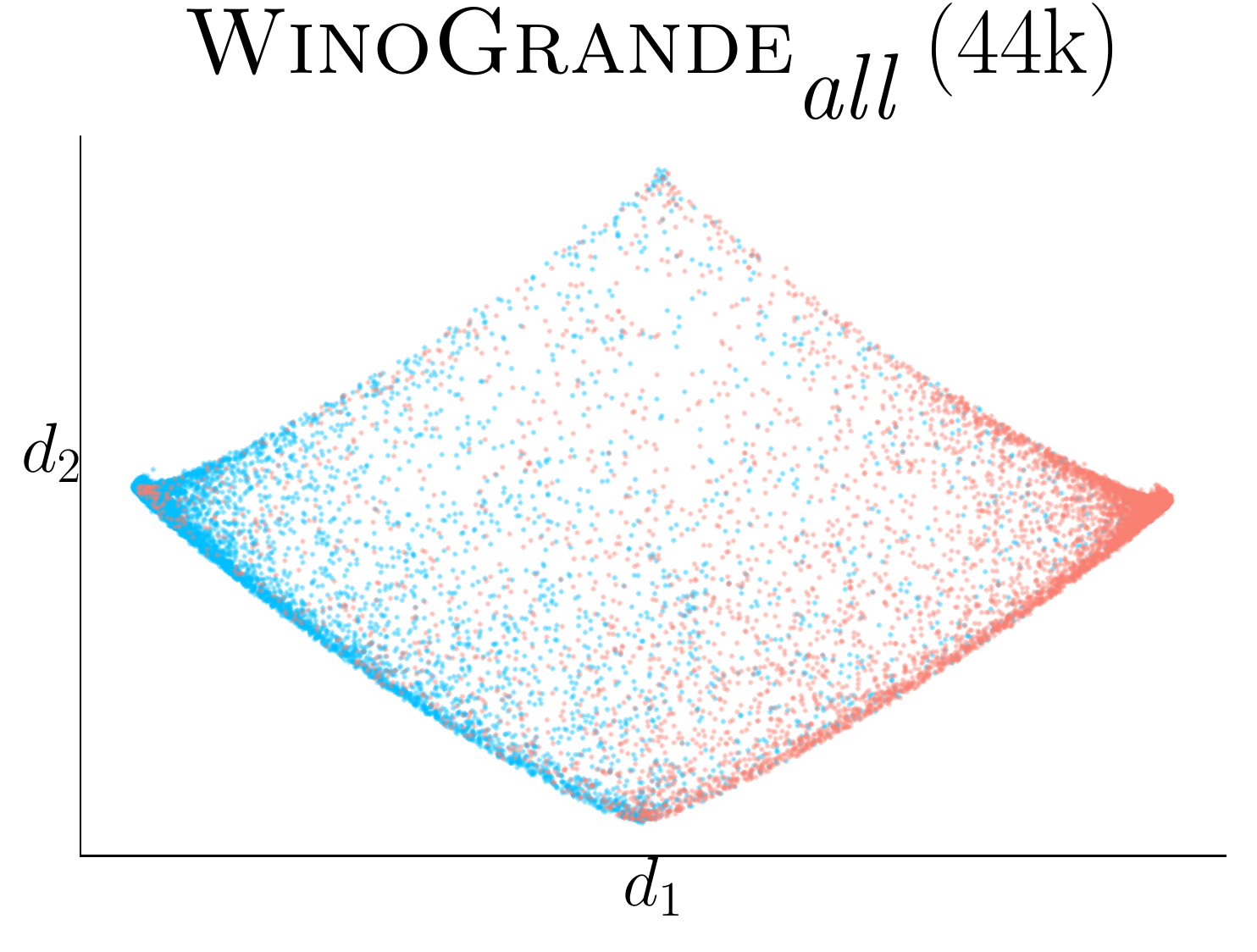}}
    \hfill
    \subfigure{\includegraphics{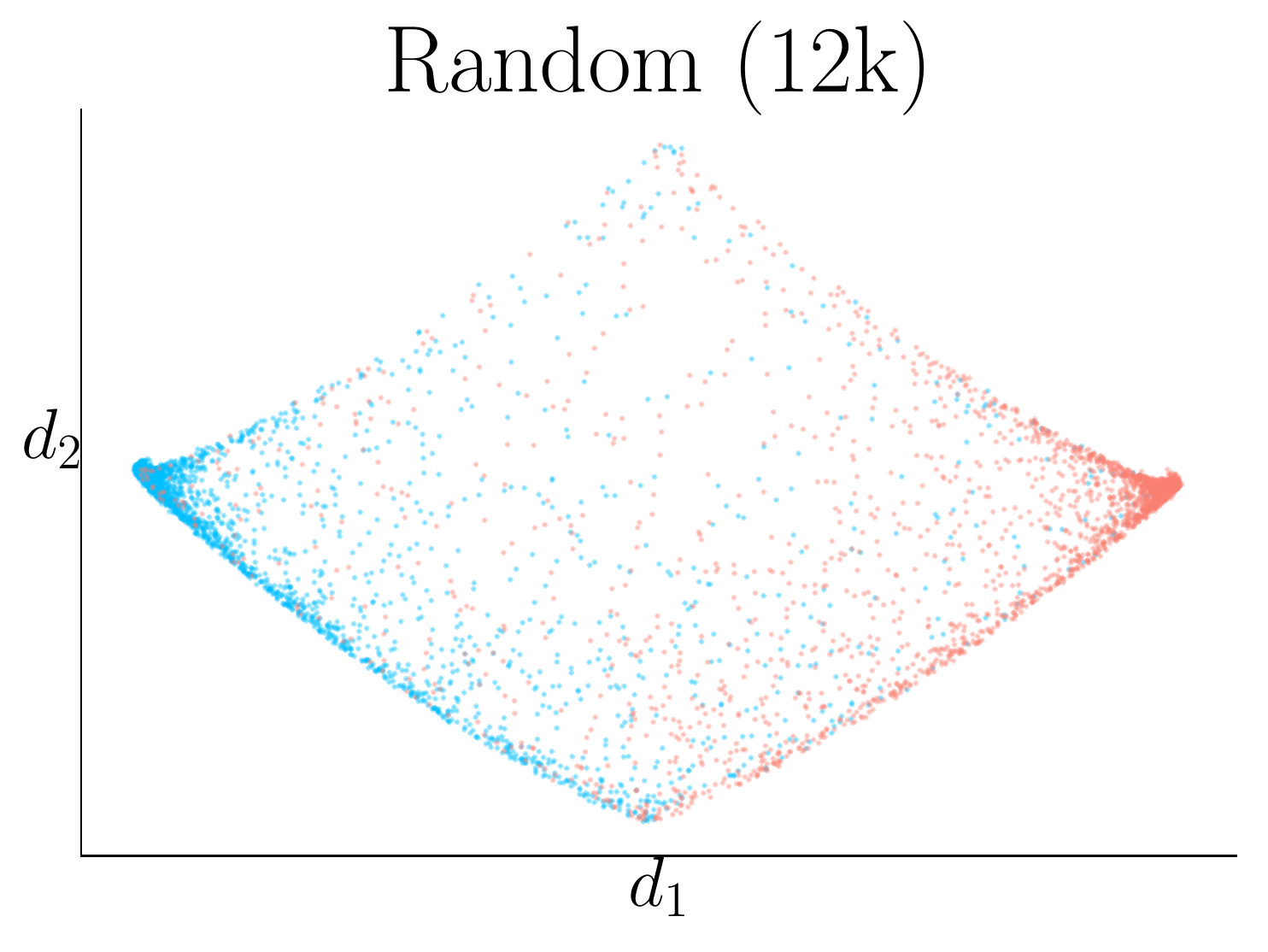}}
    \hfill
    \subfigure{\includegraphics{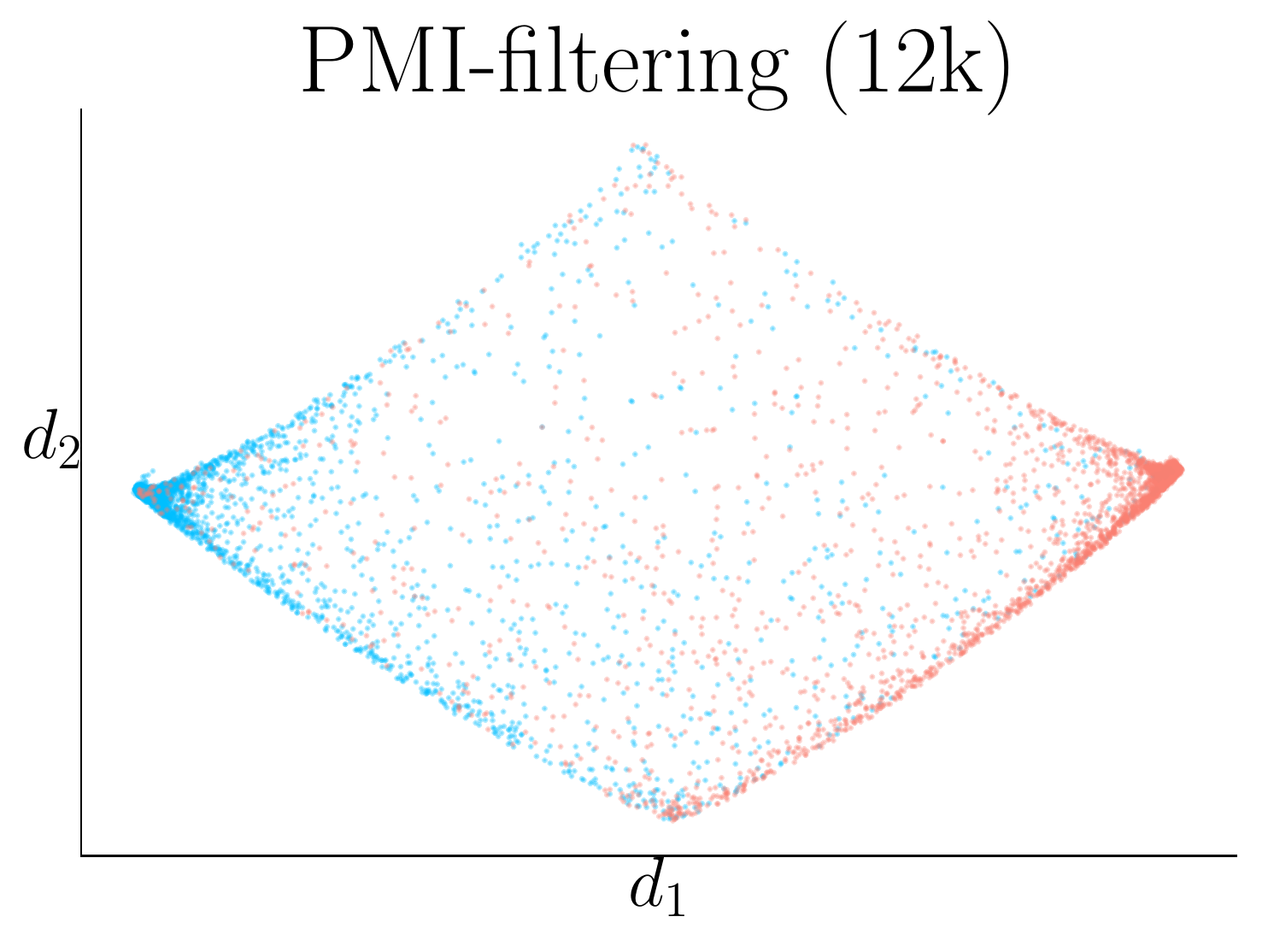}}
    \hfill
    \subfigure{\includegraphics{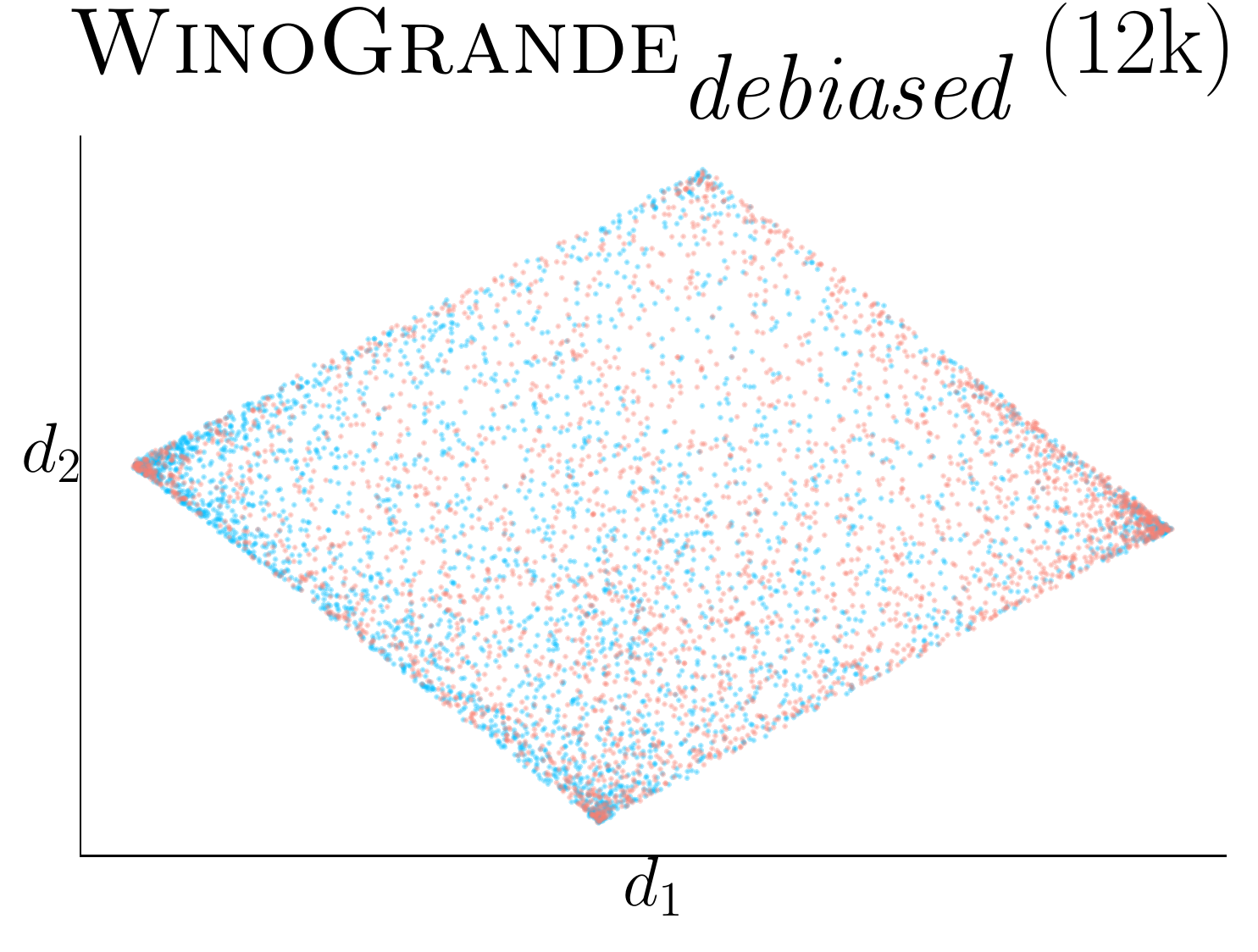}}
    \hfill
    \subfigure{\includegraphics{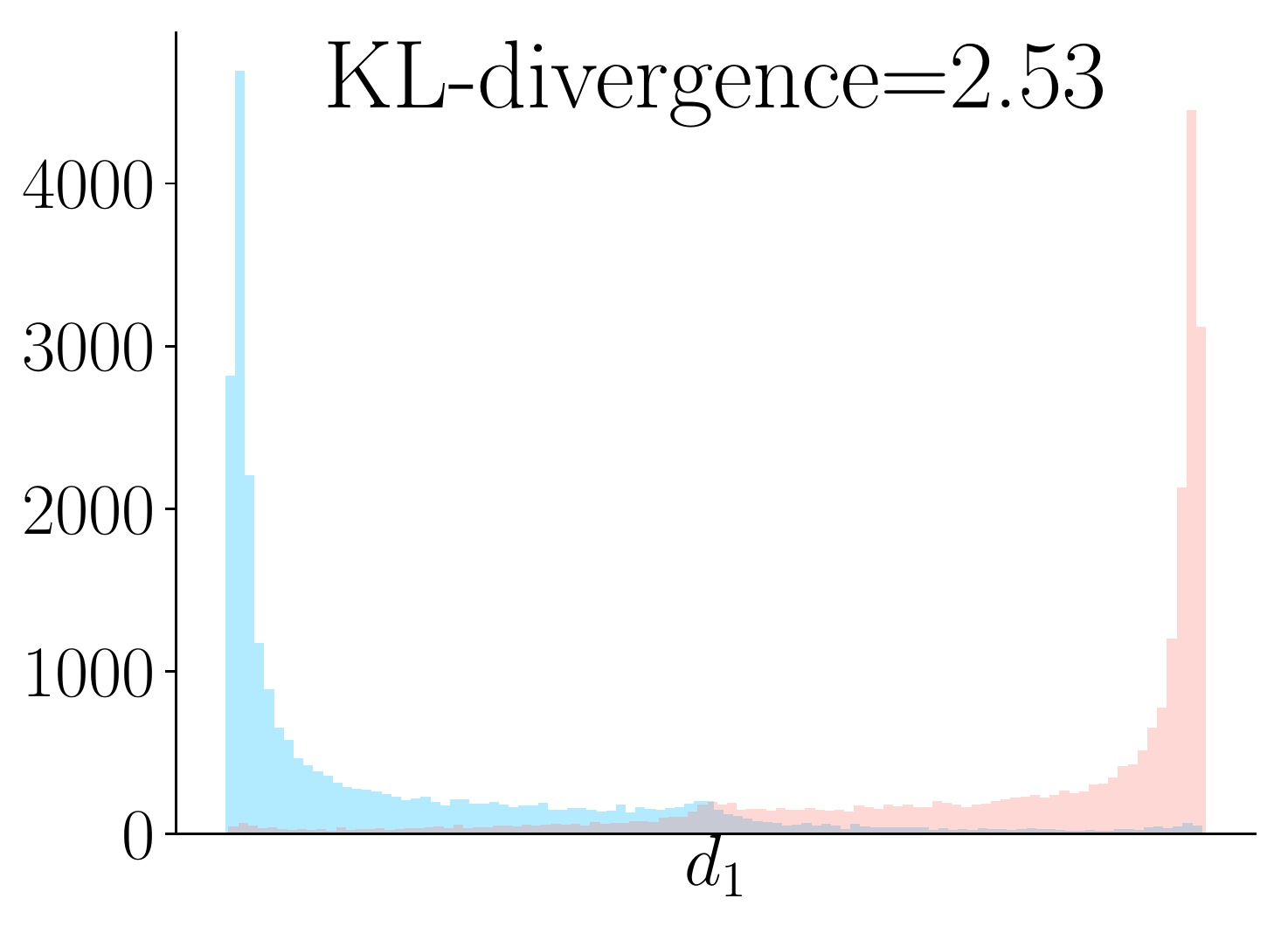}}
    \hfill
    \subfigure{\includegraphics{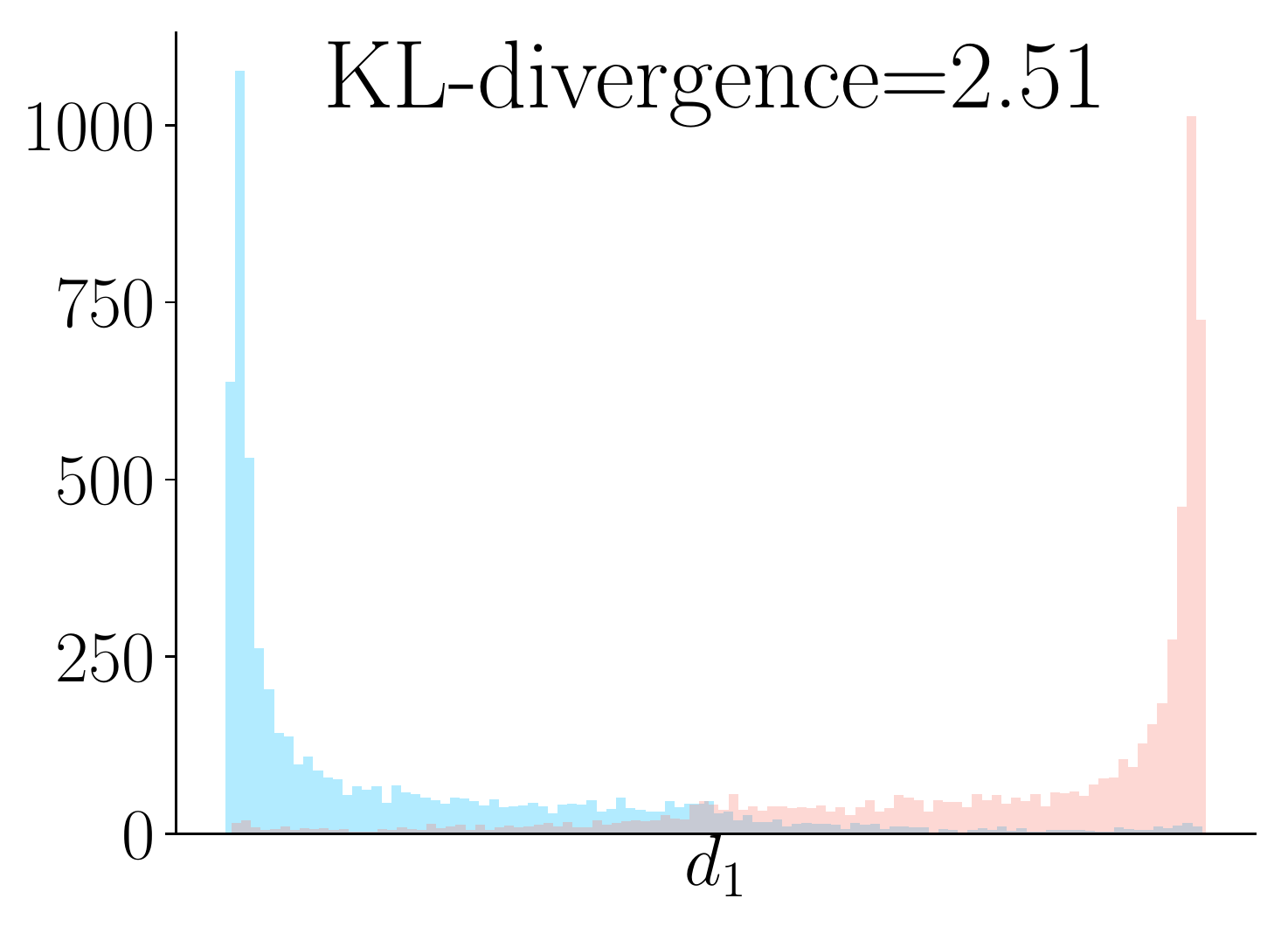}}
    \hfill
    \subfigure{\includegraphics{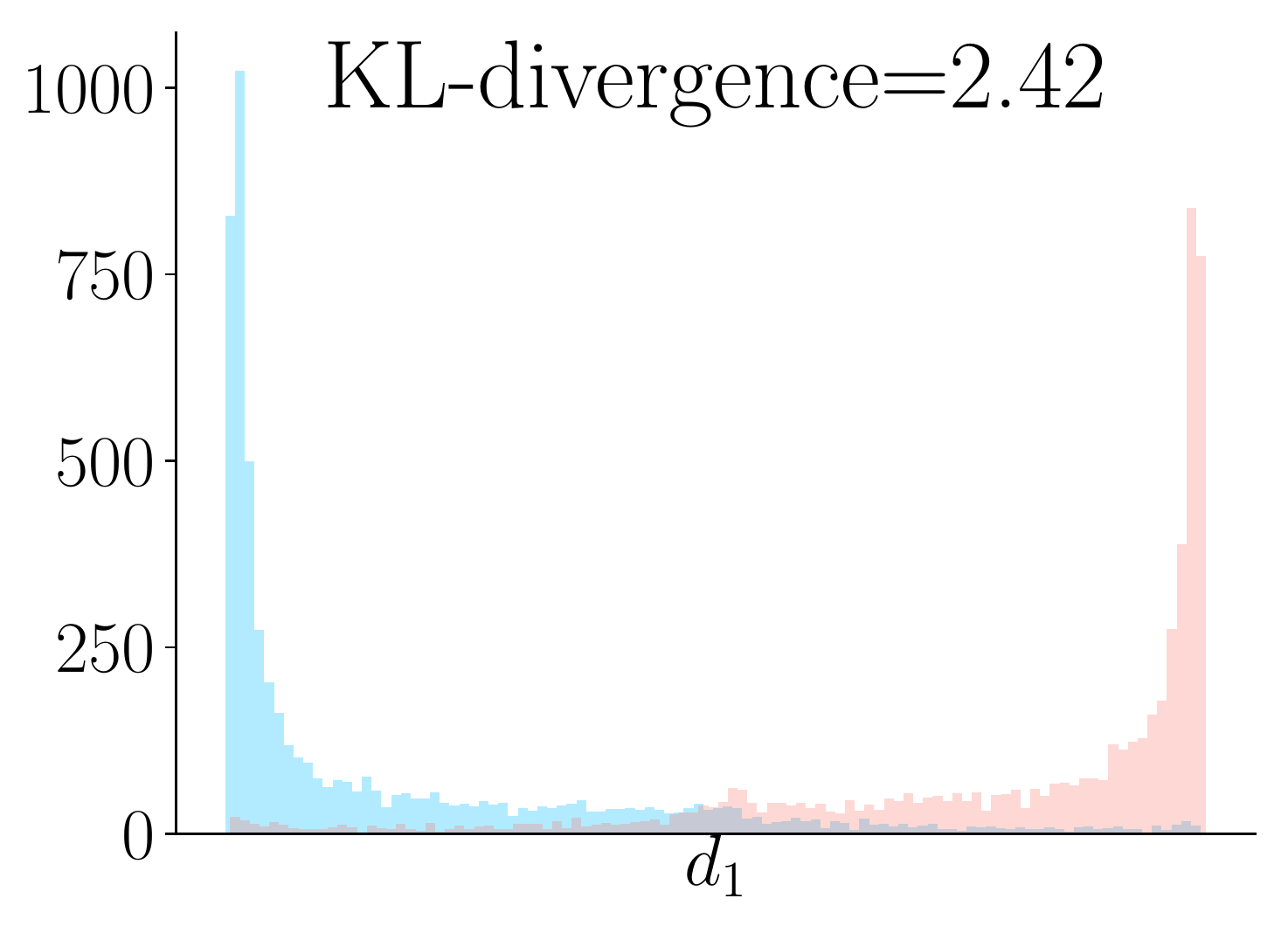}}
    \hfill
    \subfigure{\includegraphics{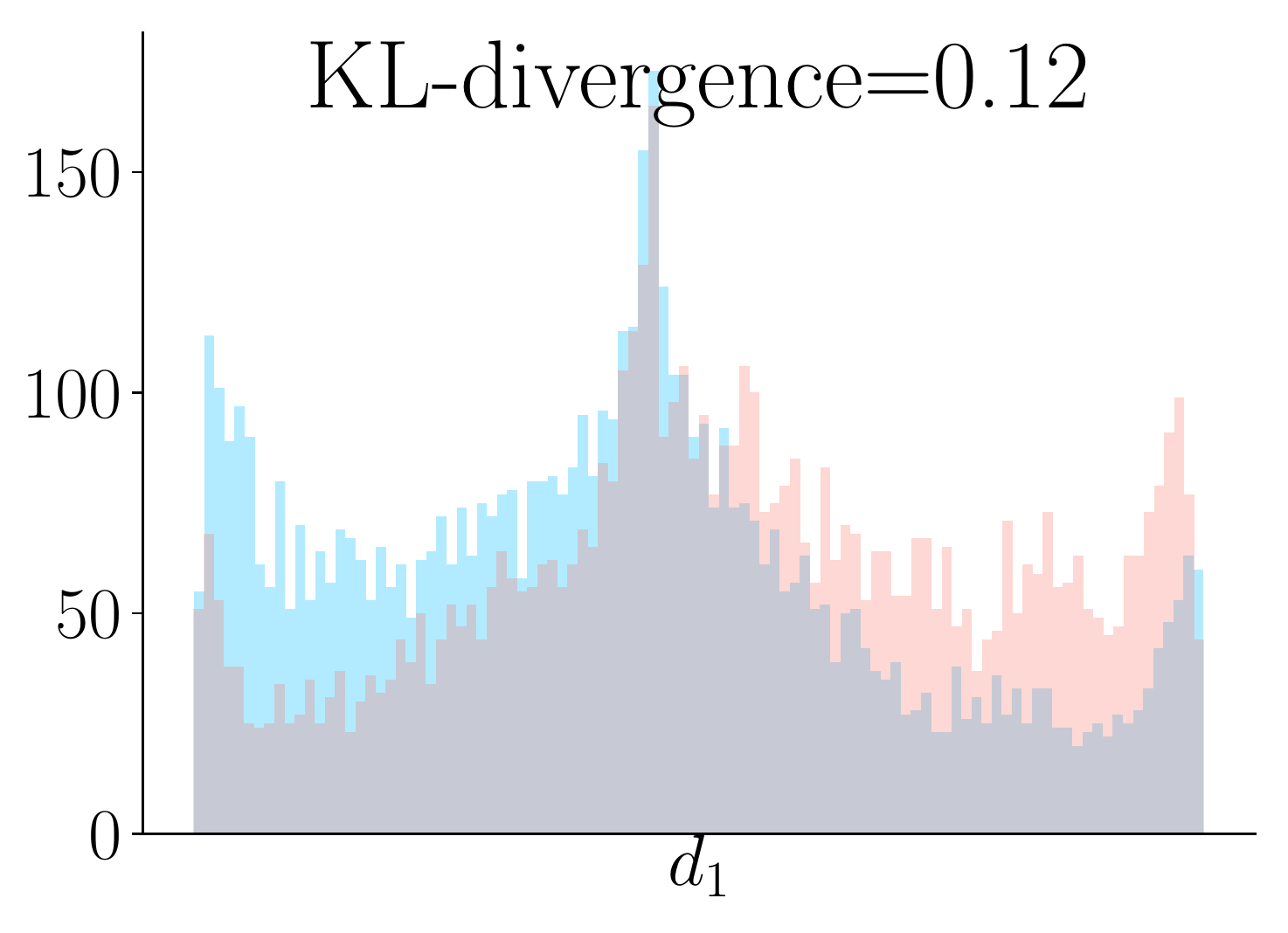}}
    \hfill
    \caption{The effect of debiasing by \algoname. RoBERTa pre-computed embeddings (applied PCA for dimension reduction) are shown in two-dimensional space (\emph{top row}) and histograms regarding $d_1$ (\emph{bottom row}) with the bin size being 100. Data points are colored depending on the label (i.e., the answer $y$ is option 1 (blue) or 2 (red)). In the histograms, we show the KL-divergence between $p(d_1, y$=1$)$ and $q(d_1, y$=2$)$.}
    \label{fig:filtering}
\end{figure*}

\begin{table*}[t]
\centering
\begin{tabular}{cll}
\hline
\multicolumn{2}{c}{\small{Twin sentences }}      & \small{Options (\textbf{answer})} \\ \hline
\multirow{2}{*}{\small{\xmark}}  & \small{The monkey \sethlcolor{teagreen}\hl{loved} to play with the balls but \sethlcolor{piggypink}\hl{ignored} the blocks because he found \textbf{them} \underline{\textit{\sethlcolor{teagreen}\hl{exciting}}}.}  & \small{\textbf{balls} / blocks} \\
                                 & \small{The monkey \sethlcolor{teagreen}\hl{loved} to play with the balls but \sethlcolor{piggypink}\hl{ignored} the blocks because he found \textbf{them} \underline{\textit{\sethlcolor{piggypink}\hl{dull}}}.}  & \small{balls / \textbf{blocks}} \\ \hline
\multirow{2}{*}{\small{\xmark}}  & \small{William could \sethlcolor{piggypink}\hl{only climb begginner} walls while Jason \sethlcolor{teagreen}\hl{climbed advanced} ones because \textbf{he} was very  \underline{\textit{\sethlcolor{piggypink}\hl{weak}}}.} & \small{\textbf{William} / Jason} \\
                                 & \small{William could \sethlcolor{piggypink}\hl{only climb begginner} walls while Jason \sethlcolor{teagreen}\hl{climbed advanced} ones because \textbf{he} was very  \underline{\textit{\sethlcolor{teagreen}\hl{strong}}}.} & \small{William / \textbf{Jason}} \\ \hline \hline
\multirow{2}{*}{\small{\cmark}}  & \small{Robert woke up at 9:00am while Samuel woke up at 6:00am, so \textbf{he} had \underline{\textit{less}} time to get ready for school.} & \small{\textbf{Robert} / Samuel} \\
                                 & \small{Robert woke up at 9:00am while Samuel woke up at 6:00am, so \textbf{he} had \underline{\textit{more}} time to get ready for school.} & \small{Robert / \textbf{Samuel}} \\ \hline
\multirow{2}{*}{\small{\cmark}}  & \small{The child was screaming after the baby bottle and toy fell. Since the child was \underline{\textit{hungry}}, \textbf{it} stopped his crying.} & \small{\textbf{baby bottle} / toy} \\
                                 & \small{The child was screaming after the baby bottle and toy fell. Since the child was \underline{\textit{full}}, \textbf{it} stopped his crying.} & \small{baby bottle / \textbf{toy}} \\ \hline
\end{tabular}
\caption{Examples that have \textit{dataset-specific} bias detected by \algoname (marked with \xmark). The words that include (dataset-specific) polarity bias (\S\ref{sec:filtering}) are highlighted ({\sethlcolor{teagreen}\hl{positive}} and {\sethlcolor{piggypink}\hl{negative}}). For comparison, we show examples selected from \datasetdeb (marked with \cmark).}
\label{tab:biased-examples}
\end{table*}

\para{Assessment of \algoname{}}
We assess the impact of \algoname relative to two baselines: random data reduction and PMI-based filtering.
In random data reduction, we randomly subsample the dataset to evaluate how a decrease in dataset size affects the bias.
In PMI-based filtering,
we compute the difference ($f$) of PMIs for each twin ($t$) as follows:
\begin{align}
f(t_1, t_2) = \sum_{w\in{t_1}}\text{PMI}(y=1;w) - \sum_{w\in{t_2}}\text{PMI}(y=1;w). \nonumber
\end{align}
Technically, we first pre-computed PMI between a word and the label $y=1$ for each word in the dataset, following a method proposed by \newcite{gururangan-etal-2018-annotation}.
The sum of PMI value of each token in a given sentence indicates the likelihood of the label $y=1$ for the sentence. 
We only retain twins that have a small difference in their PMI values as it corresponds to twins that are hard to discriminate.\footnote{We also evaluated other variations of PMI-filtering such as the absolute difference ($\lvert f\rvert$), max-PMI ($\max(\max_{w\in{t_1}}\text{PMI}(y;w), \max_{w\in{t_2}}\text{PMI}(y;w))$), and token-pairwised PMI($y; w_1, w_2\in t$), but we did not observe a significant difference among them.}

Figure~\ref{fig:filtering} plots RoBERTa pre-computed embeddings whose dimension is reduced to 2D (\emph{top}) and 1D (\emph{bottom}) using Principal Component Analysis (PCA).  
We observe that \datasetall and the two baselines exhibit distinct components between the two correct answer options (i.e., $y\in{1,2}$), whereas such distinction becomes less salient in \datasetdeb, which implies that \algoname successfully reduces the spurious correlation in the dataset (between instances and labels).
To quantify the effect, we compute the KL divergence between the samples with answer options.
We find that the random data reduction does not reduce the KL divergence ($2.53$ $\rightarrow$ $2.51$).
It is interesting to see that PMI-filtering marginally reduces the KL divergence ($\rightarrow$ $2.42$), although the principal component analysis on the PMI-filtered subset still leads to a significant separation between the labels.
On the other hand, in \datasetdeb, \algoname reduces the KL divergence dramatically ($\rightarrow$ $0.12$) which suggests that this debiased dataset should be challenging for statistical models that solely rely on spurious correlation.

\para{What bias has been actually detected by \algoname?}
Is the bias really spurious and undesirable according to the original WSC's goal?
Table~\ref{tab:biased-examples} presents examples that \algoname has detected as a dataset-specific bias.
We see a structural pattern in the first two twins, where the sentiment between the answer option and the target pronoun are highly correlated.
In other words, these problems can be easily answered by simply exploiting the pattern of the polarity (positive or negative).
Importantly, this dataset-specific bias is structural rather than at the token level, contrasting with the biases that have been identified in the NLI literature
~\cite{gururangan-etal-2018-annotation,poliak-etal-2018-hypothesis}, and it is hard to detect these biases using heuristics such as lexical PMI-filtering. Instead of depending on such heuristics, \algoname is able to detect samples that potentially have such biases algorithmically.

After applying the \algoname{} algorithm, we obtain a \textit{debiased} dataset of 12,282 instances split into training (9,248), development (1,267), and test (1,767) sets.
We also release 31k problems that are filtered out by \algoname{} for additional training set (\S\ref{sec:experiments}) and resource (\S\ref{sec:application}), resulting in a total number of problems in \datasetall to be 43,972 (40,938 for training, 1,267 for development, and 1,767 for test).

\subsection{\textsc{WinoGrande} V.S. the Original WSC}
While \dataset is inspired by the original WSC, we make a few design choices that deviate from the original design guidelines of WSC in order to scale up the dataset considerably while ensuring the hardness of the dataset. 

First, 
\dataset is formatted as a fill-in-the-blank problem where the blank corresponds to the mention of one of the two names in the context, following the same modification made by other recent WSC variants such as \newcite{trinh2018simple}.\footnote{ \url{https://github.com/tensorflow/models/tree/master/research/lm_commonsense}} In contrast, the original WSC explicitly places a pronoun (instead of a blank). From the modeling stand point, the use of blanks instead of explicit pronouns do not make the problem any easier. 

Second, while we originally collected all problems in twins, the final questions in the filtered \datasetdeb are not always twins because it is possible that \algoname filters out only one of the twin sentences. In \datasetdeb, about 1/3 of questions are not twins. We also release \datasetall (training set) that all consists of twins. 

Third, unlike the original WSC problems that were composed by just a few linguistics experts, \dataset is authored by crowdworkers. Thus, the language used in \dataset reflects the more diverse and noisy language used by crowds. Importantly, laymen still find \dataset problems easy to solve, with 94\% accuracy (\S\ref{sec:experiments}).

\section{Experimental Results}
\label{sec:experiments}

\subsection{Baseline Models}
We evaluate the \datasetdeb (dev and test) on methods/models that have been effective on the original WSC.

\para{Wino Knowledge Hunting}
Wino Knowledge Hunting (WKH) by ~\newcite{emami-etal-2018-generalized} is based on an information retrieval approach, where the sentence is parsed into a set of queries and then the model looks for evidence for each answer candidate from the search result snippets.
This IR-oriented approach comes from a line of work in coreference resolution~\cite{kobdani-etal-2011-bootstrapping,ratinov-roth-2012-learning,bansal-klein-2012-coreference,zheng-etal-2013-dynamic,peng-etal-2015-solving}.

\para{Ensemble Neural LMs}
\newcite{trinh2018simple} is one of the first attempts to apply a neural language model which is pre-trained on a very large corpora (including LM-1-Billion, CommonCrawl, SQuAD, and Gutenberg Books).
In this approach, the task is treated as fill-in-the-blank question with binary choice.
The target pronoun in the sentence is replaced by each answer candidate and the neural language model provides the likelihood of the two resulting sentences.
This simple yet effective approach outperforms previous IR-based methods.

\para{BERT}
BERT~\cite{bert} is another pre-trained neural language model which has bidirectional paths and consecutive sentence representations in hidden layers.
We finetune BERT with splitting the input sentence into context and option using the candidate answer as delimiter.
The input format becomes \texttt{[CLS] context [SEP] option [SEP]}; e.g., \textit{The trophy doesn't fit into the brown suitcase because the \blank \texttt{[SEP]} is too large. \texttt{[SEP]}} (The blank \blank is filled with either option 1 or 2), 
and the [CLS] token embedding is used to classify which answer option is correct.
We used grid-search for hyper-parameter tuning: learning rate $\{1e-5, 3e-5, 5e-5\}$, number of epochs $\{3, 4, 5, 8\}$, batch-size $\{8, 16\}$ with three different random seeds.

\para{RoBERTa}
RoBERTa~\cite{Liu2019RoBERTaAR} is an improved variant of BERT that adds more training data with larger batch sizes and training time, as well as other refinements such as dynamic masking.
RoBERTa performs consistently better than BERT across many benchmark datasets.

\para{Word association baseline}
Using BERT and RoBERTa, we also run the word association baseline (\textit{local-context-only}) to check if the dataset can be solved by language-based bias.
In this baseline, the model is trained with only local contexts ($w_{t-2:\text{EOS}}$) surrounding the blank to be filled ($w_t$) (e.g., because the \blank \texttt{[SEP]} is too large. \texttt{[SEP}]).
This is analogous to the \textit{hypothesis-only} baseline in NLI~\cite{poliak-etal-2018-hypothesis}, where the task (dataset) does not require the full context to achieve high performance.

\para{Finetuning on DPR dataset}
DPR (Definite Pronoun Resolusiton Dataset), collected by \newcite{rahman-ng-2012-resolving}, consists of 1,886 WSC style problems written by 30 undergraduate students. \newcite{KCCYL2019} have recently shown that BERT finetuned with DPR boosts the performance on WCS (72.2\% accuracy). 
As additional baselines, we finetune BERT and RoBERTa with DPR and evaluate on \dataset. 
This allows us to compare the difficulty of WSC and \dataset empirically.

\para{Human evaluation}
In addition to the methods described above, we compute human performance as the majority vote of three crowd workers for each question.

\begin{table}[t]
\centering
\begin{tabular}{lcc}
Methods         & dev acc. (\%) & test acc.(\%) \\ \toprule
WKH             & 49.4 & 49.6 \\
Ensemble LMs    & 53.0 & 50.9 \\
BERT            & 65.8 & 64.9 \\
RoBERTa            & {\bf 79.3} & {\bf 79.1} \\
\midrule
BERT (local context) & 52.5 & 51.9 \\
RoBERTa (local context) & 52.1 & 50.0 \\
\midrule
BERT-DPR$^{\star}$     & 50.2 & 51.0 \\
RoBERTa-DPR$^{\star}$ & 59.4 & 58.9 \\
\midrule
Human Perf.     & 94.1 & 94.0 \\
\end{tabular}
\caption{Performance of several baseline systems on \datasetdeb (dev and test). The star ($\star$) denotes that it is zero-shot setting (e.g., BERT-DPR$^{\star}$ is a BERT model fine-tuned with the DPR dataset and evaluated on \datasetdeb.)}
\label{tab:benchmark}
\end{table}

\begin{table}[t]
\centering
\begin{tabular}{lcc}
Training size   & dev acc. (\%) & test acc.(\%) \\ \toprule
XS (160) & 51.5 & 50.4 \\
S (640)  & 58.6 & 58.6 \\
M (2,558) & 66.9 & 67.6 \\
L (10,234) & 75.8 & 74.7 \\
XL (40,938) & 79.3 & 79.1 \\
\midrule
\end{tabular}
\caption{Performance of RoBERTa with different training sizes.}
\label{tab:training-size}
\end{table}

\subsection{Results}
Table~\ref{tab:benchmark} shows the results.
Two baselines, WKH and Ensemble LMs, only achieve chance-level performance (50\%).
The best model, RoBERTa, achieves $79.1\%$ test-set accuracy\footnote{When we use the debiased training set (9,248), both BERT and RoBERTa showed only chance level performance.}, whereas human performance achieve $94.0\%$, indicating that the \datasetdeb is still easy for humans to answer as desired.
Regarding the word association (i.e., local context) baselines, both BERT and RoBERTa achieve close to chance-level performance, illustrating that most \datasetdeb problems cannot be answered by local context only.
Finally, BERT and RoBERTa finetuned with DPR achieve chance-level to below 60\% accuracy, which contrast with the performance boosts on WSC (72\% by BERT (\newcite{KCCYL2019}) and 83\% in RoBERTa) and other existing WSC-style problems (shown in \S\ref{sec:transfer-results}).
This indicates that \datasetdeb consists of more challenging problems than WSC and existing variants.

\paragraph{Learning Curve}
\begin{figure}[t]
    \centering
    \includegraphics[width=0.95\columnwidth]{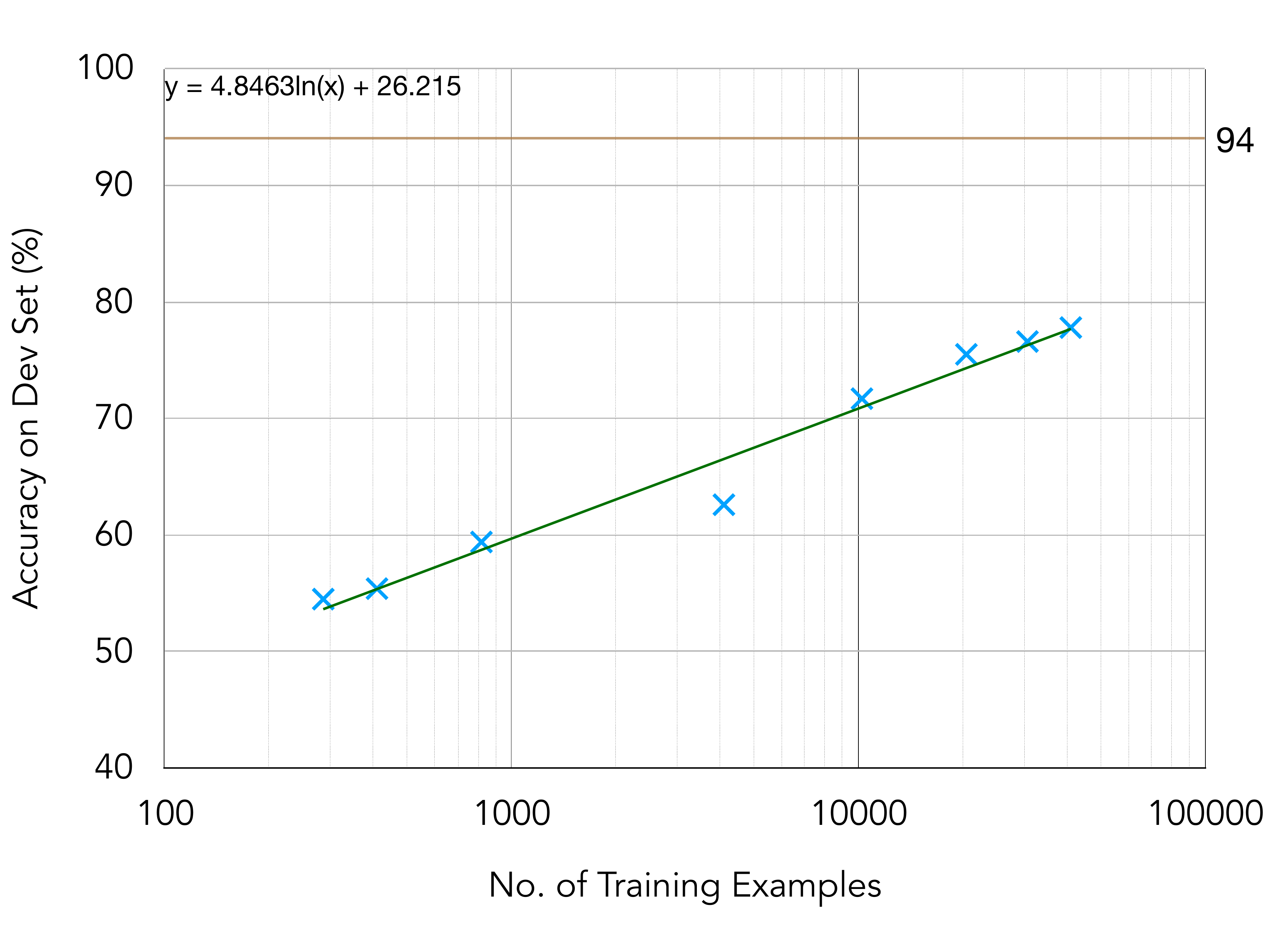}
    \caption{Learning curve on the dev set of \dataset. Each point on the plot is the best performance for a given number of randomly selected training examples, computed over ten random seeds.}
    \label{fig:lc}
\end{figure}
In order to see the effect of training size, Table~\ref{tab:training-size} shows the performance by RoBERTa trained on different training sizes from 160 to 40k questions. 
Figure \ref{fig:lc} shows the learning curve of the best model, RoBERTa, on the \datasetdeb dev set. RoBERTa's performance ranges from 59\% to 79\% when the size of training data is varied from 800 (2\% of the training data) to 41K (100\% of the training data) instances. To achieve human-level performance, current state-of-the-art models would need over 118K training instances. 

Importantly, the lower end of the available training data ($\sim$800) in the learning curve roughly matches the size of the training data made available in previous variants of WSC (see Table \ref{tab:datasets}).
For most of these datasets, state-of-the-art already reaches around 90\% (\S\ref{sec:application}). In contrast, when we control for the training set size in \dataset, RoBERTa's performance is considerably lower (59\%) -- demonstrating that our dataset construction method is able to compose WSC problems that are collectively considerably harder than previous datasets.

\section{Transfer Learning from \dataset}
\label{sec:application}
\dataset{} contains a large number of WSC style questions. In addition to serving as a benchmark dataset, we use \dataset as a resource -- we apply transfer learning by first fine-tuning a model on our dataset and evaluating its performance on related datasets: WSC, PDP, SuperGLUE-WSC, DPR, KnowRef, KnowRef, and Winogender). We establish state-of-the-art results across several of these existing benchmark datasets.

\subsection{Existing WSC and Related Datasets}
\label{sec:wsc}
We briefly describe existing WSC variants and other related datasets.
Table~\ref{tab:datasets} provides their summary statistics. 

\para{WSC~\cite{winograd}} This is the original Winograd Schema Challenge dataset, which consists of 273 problems.
The problems are manually crafted by the authors to avoid word association bias as much as possible, 
although ~\newcite{trichelair2018evaluation} later report that 13.5\% of the questions may still have word-association bias.

\para{PDP~\cite{morgenstern2016planning}}
PDP (Pronoun Disambiguation Problems) dataset is closely related to the original WSC, and used in the 2016 running of the Winograd Schema Challenge.
The dataset consists of $80$ pronoun disambiguation problems.
It is formulated as a multiple choice task, in which a pronoun must be resolved to one of up to $5$ (but mostly binary) possible antecedents.

\para{SuperGLUE-WSC~\cite{wang2019superglue}}
SuperGLUE contains multiple datasets including a modified version of WSC, which we will refer to as SuperGLUE-WSC.
This dataset aggregates the original WSC, PDP and additional PDP-style examples, and recasts them into True/False binary problems (e.g., ``Pete envies \textbf{Martin} because \textit{he} is very successful.'' Q: Does \textit{he} refer to \textbf{Martin}? A: True).
Therefore, the number of problems are roughly doubled from WSC and PDP, although the size is still relatively small (804 in total). We converted WinoGrande to the True/False binary problems.

\begin{table}[t!]
\centering
\begin{tabular}{l|r|r|r}
\hline
Dataset                         & \#Probs & Avg Len & \#Vocab \\ \hline
WSC                 & 273    & 19.1 & 919    \\
PDP                 & 80     & 39.5 & 594    \\
SuperGLUE-WSC       & 804    & 28.4 & 1,711   \\
DPR                 & 1,886  & 15.9 & 4,127  \\
KnowRef             & 1,269  & 19.3 & 5,310   \\
COPA                & 1,000  & 13.3 & 3,369  \\
Winogender          & 720    & 15.6 & 523    \\
\datasetdeb         & 12,282 & 21.1 & 11,408 \\
\datasetall         & 43,972 & 20.6 & 16,469 \\ \hline
\end{tabular}
    \caption{Statistics on WSC and related datasets (\S\ref{sec:wsc}).
    }
\label{tab:datasets}
\end{table}

\para{DPR~\cite{rahman-ng-2012-resolving}}
DPR (Definite Pronoun Resolution Dataset) introduces 1,886 additional WSC problems authored by 30 undergraduate students.
\newcite{trichelair2018evaluation} point out that this dataset is overall less challenging than the original WSC due to an increased level of language-based or dataset-specific biases.
We split the original training set (1,332) into training (1,200) and development (122) sets, DPR does not have an official split for it. 

\para{KnowRef~\cite{emami-etal-2019-knowref}}
KnowRef provides over 8k WSC-style coreference resolution problems that are extracted and filtered with heuristic rules from 100 million web sentences (Reddit, Wikipedia, and OpenSubtitles).
We report results on the publicly available \emph{test} set (1.2k problems).

\para{COPA~\cite{copa}}
This dataset introduces 1,000 problems that
aim to test commonsense reasoning focusing on script knowledge, formulated as a binary choice about \emph{causes} and \emph{effects} of given premises. 
Since COPA does not provide a training set, we split the original development set (500) into training (400) and development (100) sets in the same way as SuperGLUE-COPA~\cite{wang2019superglue}.

\para{Winogender~\cite{rudinger-etal-2018-gender}} This dataset introduces $720$ problems focusing on pronouns resolution with respect to people, with distinct goal of measuring gender bias in coreference resolution systems.

\subsection{Experimental Setup}
Our model is based on RoBERTa finetuned with \dataset{} (train and dev sets). To compare different corpora used as a resource, we also finetune RoBERTa on DPR (train and test sets).
For hyper parameter search, we use the same grid search strategy as in \S\ref{sec:experiments}.

\para{Additional Human Evaluation}
We also report human performance for WSC, PDP, and DPR to calibrate the quality of our crowd worker pool as well as to support previous findings.
To our knowledge, this is the first work to report human performance on the DPR dataset.

\begin{table}[h!]
\centering
\begin{tabular}{ll}
\hline
\multicolumn{2}{c}{WSC\small{~\cite{winograd}}} \\ \hline
\multicolumn{1}{l|}{\newcite{liu2016commonsense}}   & 52.8 \\
\multicolumn{1}{l|}{WKH\small{~\cite{emami-etal-2018-generalized}}}   & 57.1 \\
\multicolumn{1}{l|}{Ensemble LMs\small{~\cite{trinh2018simple}}} & 63.8 \\
\multicolumn{1}{l|}{GPT2\small{~\cite{radford2019language}}}  & 70.7 \\
\multicolumn{1}{l|}{BERT-DPR$^\star$\small{~\cite{KCCYL2019}}}  & 72.2 \\
\multicolumn{1}{l|}{HNN\small{~\cite{He2019AHN}}}  & 75.1$^\dagger$ \\
\multicolumn{1}{l|}{RoBERTa-DPR$^\star$ (This work)}  & 83.1 \\
\multicolumn{1}{l|}{\textbf{RoBERTa-WinoGrande$^\star$ (This work)}} & \textbf{90.1} \\ \hdashline
\multicolumn{1}{l|}{Humans\small{~\cite{bender2015establishing}}} & 92.1 \\
\multicolumn{1}{l|}{Humans (This work)}     & 96.5 \\ \hline
\hline
\multicolumn{2}{c}{PDP\small{~\cite{morgenstern2016planning}}} \\ \hline
\multicolumn{1}{l|}{\newcite{liu2016commonsense}}   & 61.7      \\
\multicolumn{1}{l|}{\newcite{trinh2018simple}} & 70.0     \\
\multicolumn{1}{l|}{RoBERTa-DPR$^\star$ (This work)}  & 86.3 \\
\multicolumn{1}{l|}{\textbf{RoBERTa-WinoGrande$^\star$ (This work)}} & 87.5 \\ 
\multicolumn{1}{l|}{HNN\small{~\cite{He2019AHN}}}  & \textbf{90.0}$^\dagger$ \\ \hdashline
\multicolumn{1}{l|}{Humans \small{~\cite{davis2016human}}}   & 90.9 \\
\multicolumn{1}{l|}{Humans (This work)}     & 92.5 \\ \hline
\hline
\multicolumn{2}{c}{SuperGLUE-WSC\small{~\cite{wang2019superglue}}}\\ \hline
\multicolumn{1}{l|}{Majority baseline} & 65.1      \\
\multicolumn{1}{l|}{RoBERTa-DPR-ft (This work)} & 83.6 \\ 
\multicolumn{1}{l|}{\textbf{RoBERTa-WinoGrande-ft (This work)}} & 85.6 \\ 
\multicolumn{1}{l|}{RoBERTa-ensemble\small{~\cite{Liu2019RoBERTaAR}}} & \textbf{89.0} \\ \hdashline
\multicolumn{1}{l|}{Humans~\small{\cite{wang2019superglue}}}     & 100 \\ \hline
\hline
\multicolumn{2}{c}{DPR\small{~\cite{rahman-ng-2012-resolving}}}\\ \hline
\multicolumn{1}{l|}{\newcite{rahman-ng-2012-resolving}}   & 73.0     \\
\multicolumn{1}{l|}{\newcite{peng-etal-2015-solving}}   & 76.4     \\
\multicolumn{1}{l|}{BERT-WinoGrande$^\star$ (This work)}  & 84.9 \\
\multicolumn{1}{l|}{RoBERTa-ft (This work)}  & 91.7 \\
\multicolumn{1}{l|}{RoBERTa-WinoGrande$^\star$ (This work)}  & 92.5 \\
\multicolumn{1}{l|}{\textbf{RoBERTa-WinoGrande-ft (This work)}} & \textbf{93.1} \\ \hdashline
\multicolumn{1}{l|}{Humans (This work)}     & 95.2 \\ \hline
\hline
\multicolumn{2}{c}{KnowRef\small{~\cite{emami-etal-2019-knowref}}}\\ \hline
\multicolumn{1}{l|}{\newcite{emami-etal-2019-knowref}} & 65.0      \\
\multicolumn{1}{l|}{RoBERTa-DPR$^\star$ (This work)}        & 84.2 \\
\multicolumn{1}{l|}{\textbf{RoBERTa-WinoGrande$^\star$ (This work)}} & \textbf{85.6}     \\ \hdashline
\multicolumn{1}{l|}{Humans~\small{\cite{emami-etal-2019-knowref}}}     & 92.0 \\ \hline
\hline
\multicolumn{2}{c}{COPA\small{~\cite{copa}}}\\ \hline
\multicolumn{1}{l|}{\newcite{Gordon2011CommonsenseCR}} & 65.4      \\
\multicolumn{1}{l|}{\newcite{Sasaki2017HandlingME}}  &  76.4     \\
\multicolumn{1}{l|}{RoBERTa-WinoGrande$^\star$ (This work)} & 84.4     \\
\multicolumn{1}{l|}{RoBERTa-ft (This work)} & 86.4$^\ddagger$ \\ 
\multicolumn{1}{l|}{\textbf{RoBERTa-WinoGrande-ft (This work)}} & \textbf{90.6}     \\ \hdashline
\multicolumn{1}{l|}{Humans~\small{\cite{gordon-etal-2012-semeval}}}     & 99.0 \\ \hline
\end{tabular}
\caption{Accuracy ($\%$) on existing WSC-related tasks (test set). The star ($\star$) denotes that it is zero-shot setting. `-ft' indicates \textit{fine-tuning} on the targeted dataset (train and dev). RoBERTa-X-ft denotes sequential fine-tuning with dataset X followed by the targeted dataset. The daggers ($\dagger$) indicate that the evaluation data is not exactly the same from ours. The double dagger ($\ddagger$) denotes that we could not reproduce the same number as in SuperGLUE leaderboard~\cite{wang2019superglue}.}
\label{tab:other-tasks}
\end{table}

\subsection{Experimental Results}
\label{sec:transfer-results}
Tables~\ref{tab:other-tasks} and \ref{tab:winogender} show results of applying transfer learning from \dataset to other WSC variants.
Overall, RoBERTa fine-tuned on \dataset{} helps improve the accuracy on all the related tasks (Table~\ref{tab:other-tasks}), and performs consistently better than when RoBERTa is fine-tuned on DPR.

While improvements on some related datasets (particularly WSC, PDP, and DPR) might seem expected, the significant improvement on COPA is not so. The COPA task -- identifying causes and effects -- is very different from that in \dataset. This significant improvement on an unrelated task indicates that \dataset can  serve as a resource for commonsense knowledge transfer.

\begin{table}[t]
\hspace{-4mm}
\small
\centering
\begin{tabular}{rccccc}
\multicolumn{6}{c}{Winogender\small{~\cite{rudinger-etal-2018-gender}}} \\ \toprule
\multicolumn{1}{l}{}   & Gotcha & Female & Male & $\lvert\Delta\text{F}\rvert$              & $\lvert\Delta\text{M}\rvert$              \\ \midrule
\multirow{2}{*}{\textsc{Rule}}   & No     & 38.3   & 51.7 & \multirow{2}{*}{28.3}  & \multirow{2}{*}{14.2}  \\
                        & Yes    & 10.0   & 37.5 &                        &                        \\ \midrule
\multirow{2}{*}{\textsc{Stats}}  & No     & 50.8   & 61.7 & \multirow{2}{*}{5.0}   & \multirow{2}{*}{21.7}  \\
                        & Yes    & 45.8   & 40.0 &                        &                        \\ \midrule
\multirow{2}{*}{\textsc{Neural}} & No     & 50.8   & 49.2 & \multirow{2}{*}{14.1}  & \multirow{2}{*}{2.5}   \\
                        & Yes    & 36.7   & 46.7 &                        &                        \\ \midrule
RoBERTa-DPR             & No     & 98.3   & 96.7 & \multirow{2}{*}{1.6}   & \multirow{2}{*}{0.9}  \\
(This work)             & Yes    & 96.7   & 95.8 &                        &                        \\ \bottomrule
RoBERTa-WG              & No     & 97.5   & 96.7 & \multirow{2}{*}{0.8}   & \multirow{2}{*}{0.8}  \\
(This work)             & Yes    & 96.7   & 97.5 &                        &                        \\ \bottomrule
                        
\end{tabular}
\caption{Accuracy ($\%$) and gender bias on Winogender dataset. ``Gotcha'' indicates whether the target gender pronoun (e.g., she) is minority in the correct answer option (e.g., doctor). $\lvert\Delta\text{F}\rvert$ and $\lvert\Delta\text{M}\rvert$ show the system performance gap between ``Gotcha'' and ``non-Gotcha'' for each gender (lower the better). The first three baselines are adopted from~\newcite{rudinger-etal-2018-gender}; \textsc{Rule} is \newcite{lee-etal-2011-stanfords}, \textsc{Stats} is \newcite{durrett-klein-2013-easy}, and \textsc{Neural} is \newcite{clark-manning-2016-deep}. }
\label{tab:winogender}
\end{table}

\para{Important Implications}
We consider that while these positive results over multiple challenging benchmarks are highly encouraging, they may need to be taken with a grain of salt.
In particular, these results might also indicate the extent to which spurious dataset biases are prevalent in existing datasets, which runs the risk of overestimating the true capabilities of machine intelligence on commonsense reasoning.

Our results and analysis indicate the importance of continued research on debiasing benchmarks and the increasing need for algorithmic approaches for systematic bias reduction, which allows for the benchmarks to evolve together with evolving state of the art.
We leave it as a future research question to further investigate how much of our improvements are due to dataset biases of the existing benchmarks as opposed to true strides in improving commonsense intelligence.

\subsection{Diagnostics for Gender Bias}
Winogender is designed as diagnostics for checking whether a model (and/or training corpora) suffers from gender bias.
The bias is measured by the difference in accuracy between the cases where the pronoun gender matches the occupation's majority gender (called ``non-gotcha'') or not (``gotcha'').
Formally, it is computed as follows :
\begin{align}
\Delta F &= \text{Acc}_{\text{(Female, Non-gotcha)}} - \text{Acc}_{\text{(Female, Gotcha)}} \nonumber \\
\Delta M &= \text{Acc}_{\text{(Male, Non-gotcha)}} - \text{Acc}_{\text{(Male, Gotcha)}} \nonumber
\end{align}
for female and male cases, respectively.

Large values of $\Delta F$ or $ \Delta M$ indicate that the model is highly gender-biased,
whereas $\lvert\Delta F\rvert = \lvert \Delta M\rvert =0$ (along with high accuracy) is the ideal scenario. In addition, if $\Delta F$ or $\Delta M$ is largely \textit{negative}, it implies that the model is biased in the other way around.

The result of the gender-bias diagnostics is shown in Table~\ref{tab:winogender}.
While we find that the RoBERTa models finetuned on \dataset{} and DPR both demonstrate very high accuracy,
the gender gap in RoBERTa-WinoGrande is smaller than RoBERTa-DPR.

\section{Conclusions}
\label{sec:conclusion}
We introduce \dataset{}, a new collection of 44k WSC-inspired problems that is significantly larger than existing variants of the WSC dataset. To create a dataset that is robust against spurious dataset-specific bias, we also present \algoname{} -- a novel light-weight adversarial filtering algorithm for systematic bias reduction. 
The resulting dataset is considerably more challenging for existing state-of-the-art models while still being trivially easy for humans.
In addition, using \dataset{} as a resource, we demonstrate effective transfer learning and achieve state-of-the-art results on several related benchmarks.

In parallel, we also emphasize the potential risk of overestimating the performance of the state-of-the-art methods on the existing commonsense benchmarks; these models might be solving the problems \emph{right} for the \emph{wrong} reasons, by relying on spurious statistical patterns (annotation artifacts). 

Our work suggests a new perspective for designing benchmarks for measuring progress in AI. Unlike past decades where the community constructed a \emph{static} benchmark dataset to work on for many years to come, we now need AI algorithms to compose challenges that are hard enough for AI, which requires \emph{dynamic} datasets that evolve together with the evolving state-of-the-art.

\section*{Acknowledgments}
We thank the anonymous reviewers, Dan Weld, Noah Smith, Luke Zettlemoyer, Hannaneh Hajishirzi, Oren Etzioni, Leora Morgenstern, Ernest Davis, Gary Marcus, and Yuling Gu for their thoughtful feedback. 
This research was supported in part by NSF (IIS-1524371, IIS-1714566), DARPA under the CwC program through the ARO (W911NF-15-1-0543), and DARPA under the MCS program through NIWC Pacific (N66001-19-2-4031).

\bibliography{aaai}

\begin{thebibliography}{}

\bibitem[\protect\citeauthoryear{Bansal and
  Klein}{2012}]{bansal-klein-2012-coreference}
Bansal, M., and Klein, D.
\newblock 2012.
\newblock Coreference semantics from web features.
\newblock {\em ACL}.

\bibitem[\protect\citeauthoryear{Belinkov \bgroup et al\mbox.\egroup
  }{2019}]{belinkov-etal-2019-adv-removal}
Belinkov, Y.; Poliak, A.; Shieber, S.; {Van Durme}, B.; and Rush, A.
\newblock 2019.
\newblock On adversarial removal of hypothesis-only bias in natural language
  inference.
\newblock {\em *{SEM}}.

\bibitem[\protect\citeauthoryear{Bender}{2015}]{bender2015establishing}
Bender, D.
\newblock 2015.
\newblock Establishing a human baseline for the winograd schema challenge.
\newblock {\em MAICS}.

\bibitem[\protect\citeauthoryear{Chen and Cardie}{2018}]{Chen2018MultinomialAN}
Chen, X., and Cardie, C.
\newblock 2018.
\newblock Multinomial adversarial networks for multi-domain text
  classification.
\newblock {\em NAACL}.

\bibitem[\protect\citeauthoryear{Clark and
  Manning}{2016}]{clark-manning-2016-deep}
Clark, K., and Manning, C.~D.
\newblock 2016.
\newblock Deep reinforcement learning for mention-ranking coreference models.
\newblock {\em EMNLP}.

\bibitem[\protect\citeauthoryear{Davis, Morgenstern, and
  Ortiz}{2016}]{davis2016human}
Davis, E.; Morgenstern, L.; and Ortiz, C.
\newblock 2016.
\newblock Human tests of materials for the winograd schema challenge 2016.

\bibitem[\protect\citeauthoryear{Devlin \bgroup et al\mbox.\egroup
  }{2018}]{bert}
Devlin, J.; Chang, M.-W.; Lee, K.; and Toutanova, K.
\newblock 2018.
\newblock {BERT}: Pre-training of deep bidirectional transformers for language
  understanding.
\newblock {\em arXiv:1810.04805}.

\bibitem[\protect\citeauthoryear{Durrett and
  Klein}{2013}]{durrett-klein-2013-easy}
Durrett, G., and Klein, D.
\newblock 2013.
\newblock Easy victories and uphill battles in coreference resolution.
\newblock {\em EMNLP}.

\bibitem[\protect\citeauthoryear{Elazar and
  Goldberg}{2018}]{elazar-goldberg-2018-adversarial}
Elazar, Y., and Goldberg, Y.
\newblock 2018.
\newblock Adversarial removal of demographic attributes from text data.
\newblock {\em EMNLP}.

\bibitem[\protect\citeauthoryear{Emami \bgroup et al\mbox.\egroup
  }{2018}]{emami-etal-2018-generalized}
Emami, A.; Trischler, A.; Suleman, K.; and Cheung, J. C.~K.
\newblock 2018.
\newblock A generalized knowledge hunting framework for the winograd schema
  challenge.
\newblock {\em NAACL: SRW}.

\bibitem[\protect\citeauthoryear{Emami \bgroup et al\mbox.\egroup
  }{2019}]{emami-etal-2019-knowref}
Emami, A.; Trichelair, P.; Trischler, A.; Suleman, K.; Schulz, H.; and Cheung,
  J. C.~K.
\newblock 2019.
\newblock The {K}now{R}ef coreference corpus: Removing gender and number cues
  for difficult pronominal anaphora resolution.
\newblock {\em ACL}.

\bibitem[\protect\citeauthoryear{Geva, Goldberg, and Berant}{2019}]{geva2019we}
Geva, M.; Goldberg, Y.; and Berant, J.
\newblock 2019.
\newblock Are we modeling the task or the annotator? an investigation of
  annotator bias in natural language understanding datasets.
\newblock {\em arXiv:1908.07898}.

\bibitem[\protect\citeauthoryear{Gordon, Bejan, and
  Sagae}{2011}]{Gordon2011CommonsenseCR}
Gordon, A.~S.; Bejan, C.~A.; and Sagae, K.
\newblock 2011.
\newblock Commonsense causal reasoning using millions of personal stories.
\newblock {\em AAAI}.

\bibitem[\protect\citeauthoryear{Gordon, Kozareva, and
  Roemmele}{2012}]{gordon-etal-2012-semeval}
Gordon, A.; Kozareva, Z.; and Roemmele, M.
\newblock 2012.
\newblock {S}em{E}val-2012 task 7: Choice of plausible alternatives: An
  evaluation of commonsense causal reasoning.
\newblock {\em *{SEM}}.

\bibitem[\protect\citeauthoryear{Gururangan \bgroup et al\mbox.\egroup
  }{2018}]{gururangan-etal-2018-annotation}
Gururangan, S.; Swayamdipta, S.; Levy, O.; Schwartz, R.; Bowman, S.; and Smith,
  N.~A.
\newblock 2018.
\newblock Annotation artifacts in natural language inference data.
\newblock {\em NAACL}.

\bibitem[\protect\citeauthoryear{He \bgroup et al\mbox.\egroup
  }{2019}]{He2019AHN}
He, P.; Liu, X.; Chen, W.; and Gao, J.
\newblock 2019.
\newblock A hybrid neural network model for commonsense reasoning.
\newblock {\em arXiv:1907.11983}.

\bibitem[\protect\citeauthoryear{Kobdani \bgroup et al\mbox.\egroup
  }{2011}]{kobdani-etal-2011-bootstrapping}
Kobdani, H.; Schuetze, H.; Schiehlen, M.; and Kamp, H.
\newblock 2011.
\newblock Bootstrapping coreference resolution using word associations.
\newblock {\em ACL}.

\bibitem[\protect\citeauthoryear{Kocijan \bgroup et al\mbox.\egroup
  }{2019}]{KCCYL2019}
Kocijan, V.; Cretu, A.-M.; Camburu, O.-M.; Yordanov, Y.; and Lukasiewicz, T.
\newblock 2019.
\newblock A surprisingly robust trick for the winograd schema challenge.
\newblock {\em ACL}.

\bibitem[\protect\citeauthoryear{Lee \bgroup et al\mbox.\egroup
  }{2011}]{lee-etal-2011-stanfords}
Lee, H.; Peirsman, Y.; Chang, A.; Chambers, N.; Surdeanu, M.; and Jurafsky, D.
\newblock 2011.
\newblock {S}tanford{'}s multi-pass sieve coreference resolution system at the
  {C}o{NLL}-2011 shared task.
\newblock {\em CoNLL: Shared Task}.

\bibitem[\protect\citeauthoryear{Levesque, Davis, and
  Morgenstern}{2011}]{winograd}
Levesque, H.~J.; Davis, E.; and Morgenstern, L.
\newblock 2011.
\newblock The winograd schema challenge.
\newblock {\em {AAAI Spring Symposium: Logical Formalizations of Commonsense
  Reasoning}}.

\bibitem[\protect\citeauthoryear{Liu \bgroup et al\mbox.\egroup
  }{2016}]{liu2016commonsense}
Liu, Q.; Jiang, H.; Ling, Z.-H.; Zhu, X.; Wei, S.; and Hu, Y.
\newblock 2016.
\newblock Commonsense knowledge enhanced embeddings for solving pronoun
  disambiguation problems in winograd schema challenge.
\newblock {\em arXiv:1611.04146}.

\bibitem[\protect\citeauthoryear{Liu \bgroup et al\mbox.\egroup
  }{2019}]{Liu2019RoBERTaAR}
Liu, Y.; Ott, M.; Goyal, N.; Du, J.; Joshi, M.~S.; Chen, D.; Levy, O.; Lewis,
  M.; Zettlemoyer, L.~S.; and Stoyanov, V.
\newblock 2019.
\newblock Roberta: A robustly optimized bert pretraining approach.
\newblock {\em ArXiv} abs/1907.11692.

\bibitem[\protect\citeauthoryear{Morgenstern, Davis, and
  Ortiz}{2016}]{morgenstern2016planning}
Morgenstern, L.; Davis, E.; and Ortiz, C.~L.
\newblock 2016.
\newblock Planning, executing, and evaluating the winograd schema challenge.
\newblock {\em AI Magazine} 37(1):50--54.

\bibitem[\protect\citeauthoryear{Niven and Kao}{2019}]{niven-kao-2019-probing}
Niven, T., and Kao, H.-Y.
\newblock 2019.
\newblock Probing neural network comprehension of natural language arguments.
\newblock {\em ACL}.

\bibitem[\protect\citeauthoryear{Peng, Khashabi, and
  Roth}{2015}]{peng-etal-2015-solving}
Peng, H.; Khashabi, D.; and Roth, D.
\newblock 2015.
\newblock Solving hard coreference problems.
\newblock {\em NAACL}.

\bibitem[\protect\citeauthoryear{Poliak \bgroup et al\mbox.\egroup
  }{2018}]{poliak-etal-2018-hypothesis}
Poliak, A.; Naradowsky, J.; Haldar, A.; Rudinger, R.; and Van~Durme, B.
\newblock 2018.
\newblock Hypothesis only baselines in natural language inference.
\newblock {\em *{SEM}}.

\bibitem[\protect\citeauthoryear{Radford \bgroup et al\mbox.\egroup
  }{2019}]{radford2019language}
Radford, A.; Wu, J.; Child, R.; Luan, D.; Amodei, D.; and Sutskever, I.
\newblock 2019.
\newblock Language models are unsupervised multitask learners.
\newblock {\em OpenAI Blog}.

\bibitem[\protect\citeauthoryear{Rahman and
  Ng}{2012}]{rahman-ng-2012-resolving}
Rahman, A., and Ng, V.
\newblock 2012.
\newblock Resolving complex cases of definite pronouns: The winograd schema
  challenge.
\newblock {\em EMNLP-CoNLL}.

\bibitem[\protect\citeauthoryear{Ratinov and
  Roth}{2012}]{ratinov-roth-2012-learning}
Ratinov, L., and Roth, D.
\newblock 2012.
\newblock Learning-based multi-sieve co-reference resolution with knowledge.
\newblock {\em EMNLP-CoNLL}.

\bibitem[\protect\citeauthoryear{Roemmele, Bejan, and Gordon}{2011}]{copa}
Roemmele, M.; Bejan, C.~A.; and Gordon, A.~S.
\newblock 2011.
\newblock Choice of plausible alternatives: An evaluation of commonsense causal
  reasoning.
\newblock {\em AAAI Spring Symposium: Logical Formalizations of Commonsense
  Reasoning}.

\bibitem[\protect\citeauthoryear{Rudinger \bgroup et al\mbox.\egroup
  }{2018}]{rudinger-etal-2018-gender}
Rudinger, R.; Naradowsky, J.; Leonard, B.; and Van~Durme, B.
\newblock 2018.
\newblock Gender bias in coreference resolution.
\newblock {\em NAACL}.

\bibitem[\protect\citeauthoryear{Sasaki \bgroup et al\mbox.\egroup
  }{2017}]{Sasaki2017HandlingME}
Sasaki, S.; Takase, S.; Inoue, N.; Okazaki, N.; and Inui, K.
\newblock 2017.
\newblock Handling multiword expressions in causality estimation.
\newblock {\em IWCS}.

\bibitem[\protect\citeauthoryear{Stokes}{2005}]{stokes2005creativity}
Stokes, P.~D.
\newblock 2005.
\newblock {\em Creativity from constraints: The psychology of breakthrough}.
\newblock Springer Publishing Company.

\bibitem[\protect\citeauthoryear{Trichelair \bgroup et al\mbox.\egroup
  }{2018}]{trichelair2018evaluation}
Trichelair, P.; Emami, A.; Cheung, J. C.~K.; Trischler, A.; Suleman, K.; and
  Diaz, F.
\newblock 2018.
\newblock On the evaluation of common-sense reasoning in natural language
  understanding.
\newblock {\em arXiv:1811.01778}.

\bibitem[\protect\citeauthoryear{Trinh and Le}{2018}]{trinh2018simple}
Trinh, T.~H., and Le, Q.~V.
\newblock 2018.
\newblock A simple method for commonsense reasoning.
\newblock {\em arXiv:1806.02847}.

\bibitem[\protect\citeauthoryear{Tsuchiya}{2018}]{tsuchiya-2018-performance}
Tsuchiya, M.
\newblock 2018.
\newblock Performance impact caused by hidden bias of training data for
  recognizing textual entailment.
\newblock {\em LREC}.

\bibitem[\protect\citeauthoryear{Turing}{1950}]{turing1950computing}
Turing, A.~M.
\newblock 1950.
\newblock Computing machinery and intelligence.
\newblock {\em Mind}.

\bibitem[\protect\citeauthoryear{Wang \bgroup et al\mbox.\egroup
  }{2019}]{wang2019superglue}
Wang, A.; Pruksachatkun, Y.; Nangia, N.; Singh, A.; Michael, J.; Hill, F.;
  Levy, O.; and Bowman, S.~R.
\newblock 2019.
\newblock Superglue: A stickier benchmark for general-purpose language
  understanding systems.
\newblock {\em arXiv:1905.00537}.

\bibitem[\protect\citeauthoryear{Zellers \bgroup et al\mbox.\egroup
  }{2018}]{Zellers2018SWAGAL}
Zellers, R.; Bisk, Y.; Schwartz, R.; and Choi, Y.
\newblock 2018.
\newblock Swag: A large-scale adversarial dataset for grounded commonsense
  inference.
\newblock {\em EMNLP}.

\bibitem[\protect\citeauthoryear{Zheng \bgroup et al\mbox.\egroup
  }{2013}]{zheng-etal-2013-dynamic}
Zheng, J.; Vilnis, L.; Singh, S.; Choi, J.~D.; and McCallum, A.
\newblock 2013.
\newblock Dynamic knowledge-base alignment for coreference resolution.
\newblock {\em CoNLL}.

\end{thebibliography}
\bibliographystyle{aaai}

\end{document}